\documentclass{article}

\usepackage{microtype}
\usepackage{graphicx}
\usepackage{subcaption}
\usepackage{booktabs} 
\usepackage{xcolor} 
\usepackage{hyperref}

\usepackage[preprint]{icml2026}

\usepackage{amsmath}
\usepackage{amssymb}
\usepackage{mathtools}
\usepackage{amsthm}

\usepackage[capitalize,noabbrev]{cleveref}

\theoremstyle{plain}

\theoremstyle{definition}

\theoremstyle{remark}

\newcommand{\RR}{\mathbb{R}}
\usepackage[textsize=tiny]{todonotes}

\icmltitlerunning{Transport, Don't Generate}

\begin{document}

\twocolumn[
\icmltitle{Transport, Don't Generate: Deterministic Geometric Flows for Combinatorial Optimization}


\begin{icmlauthorlist}
\icmlauthor{Benjy Friedmann}{tech}
\icmlauthor{Nadav Dym}{tech}
\end{icmlauthorlist}

\icmlaffiliation{tech}{Faculty of Mathematics, Department of Applied Mathematics, Technion - Israel Institute of Technology, Haifa, Israel}

\icmlcorrespondingauthor{Benjy Friedmann}{benjamin.fri@campus.technion.ac.il}

\icmlkeywords{Geometric Deep Learning, Flow Matching, TSP, Optimal Transport}

\vskip 0.3in
]

\printAffiliationsAndNotice{} 

\begin{abstract}
Recent advances in Neural Combinatorial Optimization (NCO) have been dominated by diffusion models that treat the Euclidean Traveling Salesman Problem (TSP) as a stochastic $N \times N$ heatmap generation task. In this paper, we propose \textbf{CycFlow}, a framework that replaces iterative edge denoising with deterministic point transport. CycFlow learns an instance-conditioned vector field that continuously transports input 2D coordinates to a canonical circular arrangement, where the optimal tour is recovered from this $2N$ dimensional representation via angular sorting. By leveraging data-dependent flow matching, we bypass the quadratic bottleneck of edge scoring in favor of \textbf{linear coordinate dynamics}. This paradigm shift accelerates solving speed by \textbf{up to three orders of magnitude} compared to state-of-the-art diffusion baselines, while maintaining competitive optimality gaps.
\end{abstract}

\section{Introduction}

Euclidean routing problems, such as the Traveling Salesman Problem (TSP), are central to Operations Research, with diverse applications in logistics, circuit design, and resource allocation. TSP is an NP-hard combinatorial optimization problem \cite{karp1972reducibility}. It can be solved to optimality using time-consuming branch-and-cut solvers like Concorde \cite{applegate2009certification}. Alternatively, fast but approximate solutions can be obtained via heuristics like 2-OPT \cite{lin1973effective} and the Christofides \cite{christofides1976worst} algorithm.

\begin{figure}[t]
    \centering
    \includegraphics[width=\linewidth]{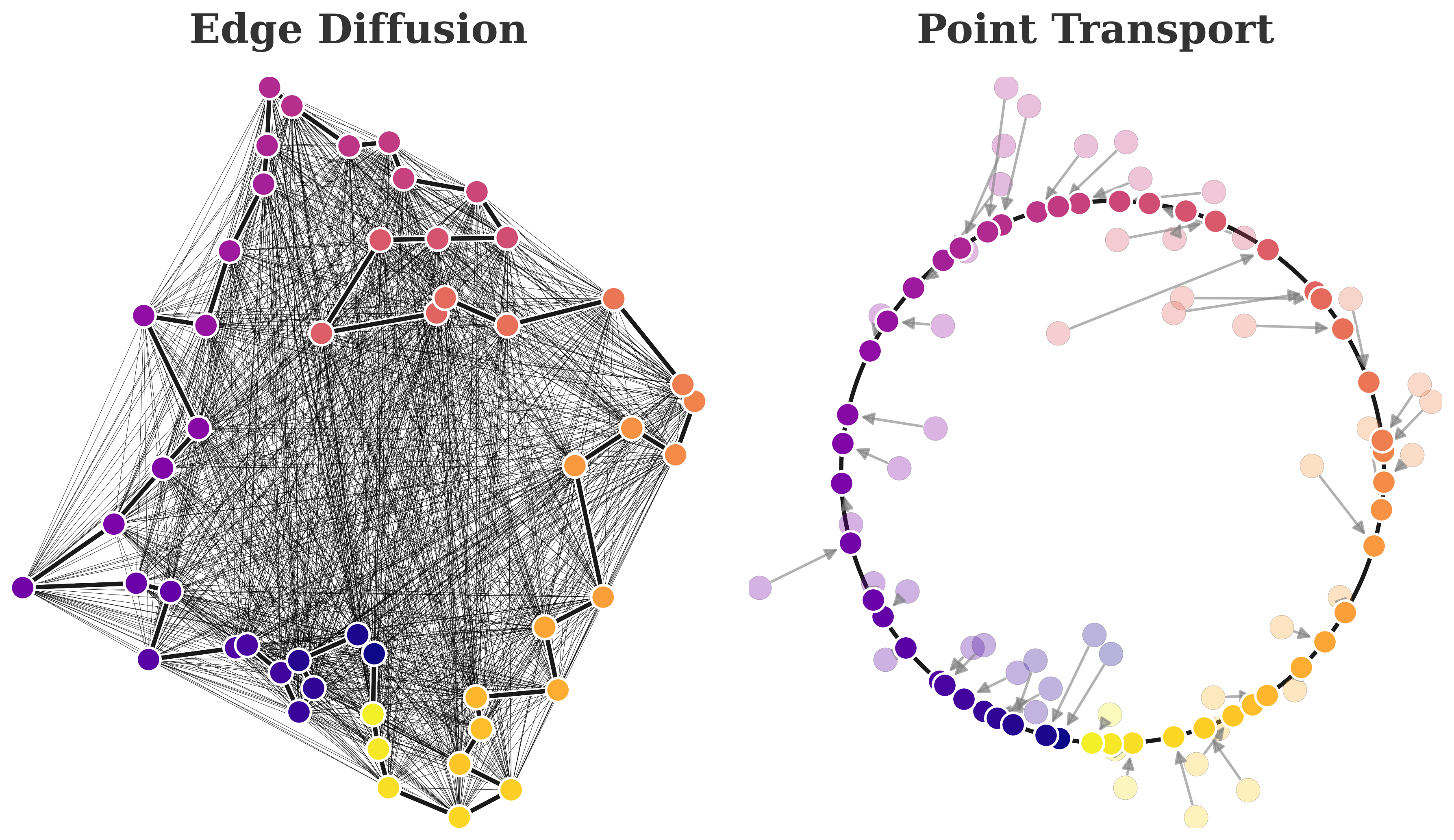}
    \caption{Comparison of the prevailing NCO paradigm (Left), which views TSP as stochastic heatmap edge denoising, versus CycFlow (Right), which treats TSP as a deterministic geometric flow. Our method transports points to a target manifold rather than classifying edges, accelerating inference by orders of magnitude.}
    \label{fig:paradigm_shift}
\end{figure}

The emergence of Neural Combinatorial Optimization (NCO) represents a paradigm shift toward data-driven solvers, where modern neural architectures learn to exploit the underlying structure of problem instances to optimize the accuracy-speed tradeoff. Initially, NCO has relied on autoregressive heuristics or constructive policies. Recently, the field has gravitated towards diffusion models, which approach TSP as a heatmap generation task. While these methods achieve strong performance by leveraging parallel denoising, they introduce a significant computational overhead: they replace sequential decoding with iterative Langevin dynamics, requiring multiple refinement steps to resolve high-fidelity heatmaps.


\begin{figure*}[t]

    \centering

    \includegraphics[width=\linewidth]{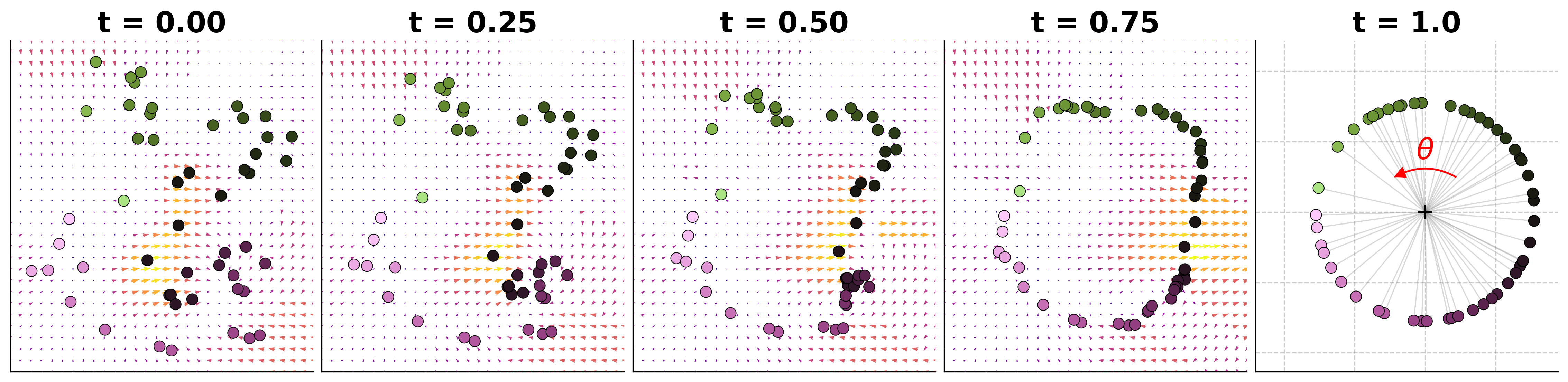}

    \caption{\textbf{Visualizing the Deterministic Linear Flow.} The panels illustrate the evolution of point sets from the initial random configuration $x_0$ ($t=0$) to the target solution manifold $x_1$ ($t=1$). The overlaid vector field indicates the flow direction $x_1 - x_0$, guiding the nodes into a structured circle. As shown in the final panel, the optimal node permutation is recovered by sorting the transported points based on their angular position $\theta$ relative to the origin.}

    \label{fig:flow_matching_vis}

\end{figure*}


 This inefficiency is not an artifact of implementation, but a direct consequence of treating a deterministic problem as a stochastic one. The TSP is inherently a deterministic geometric mapping: for a given input set, the optimal cycle is a unique well-defined structure determined entirely by the metric geometry. Formulating TSP as a generative task—sampling from a highly uncertain distribution $p(\pi|X)$—forces the solver to employ stochastic denoising, an approach that is misaligned with the problem's nature. Current state-of-the-art methods effectively sacrifice computational efficiency for distribution estimation, solving a deterministic optimization problem with probabilistic machinery designed for high-variance data generation.

In this work, we introduce CycFlow, a framework that views TSP through the lens of Deterministic Point Transport. Rather than generating a solution from Gaussian noise, CycFlow learns to transport the unordered set of nodes $X \subset \mathbb{R}^2$ onto a canonical target manifold—specifically $S^1$—where the ordering becomes trivial (\cref{fig:flow_matching_vis}). By coupling the input distribution directly to this solution manifold, our method induces flow trajectories that are straighter and simpler than those in diffusion processes. Crucially, this design shifts the computational burden from denoising $N \times N$ edge probabilities to evolving $N$ coordinate points, enabling linear-complexity inference. \cref{fig:paradigm_shift} illustrates the difference between the two approaches.

\subsection*{Contributions}
\begin{itemize}
    \item[(i)] \textbf{Transport, Not Classification:} We reformulate TSP as evolving a set of $N$ coordinates in $\RR^2$ rather than classifying an $N \times N$ adjacency matrix. This leverages the Euclidean inductive bias, treating the problem as a continuous geometric transformation rather than a discrete graph search.
    \item[(ii)] \textbf{Deterministic Flow Matching:} Utilizing Flow Matching, we regress the vector field between the input metric space and the solution geometry. This results in straight, deterministic flow paths that are easier to learn than diffusion reverse processes.
    \item[(iii)] \textbf{Orders-of-Magnitude Speedup:} We empirically demonstrate that CycFlow attains competitive  optimality gaps  while running 2-3 orders of magnitude faster than leading diffusion models, validating the efficiency of linear coordinate dynamics.
\end{itemize}

\section{Related Work}

\begin{figure*}[t]
    \centering
    \includegraphics[width=\textwidth]{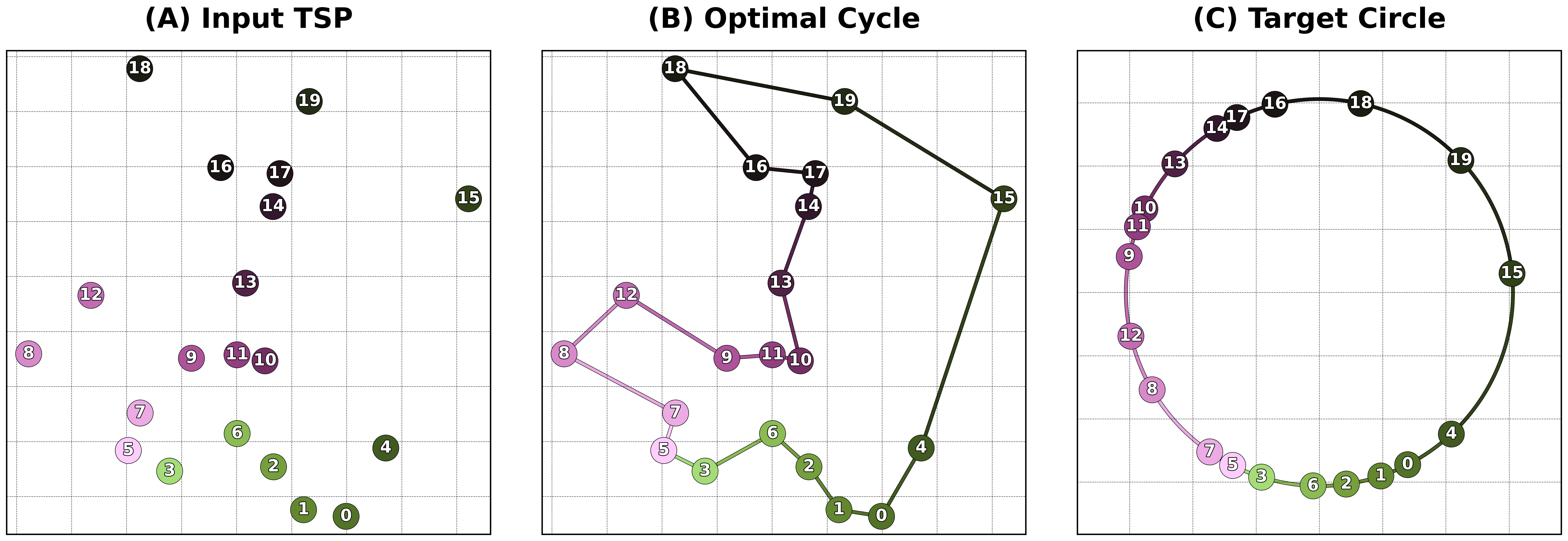}
    \caption{\textbf{Geometric Coupling.} We construct a specific target $Y$ for each input $X$ based on the optimal tour, creating a deterministic, data-dependent coupling $(x_0, x_1)$.}
    \label{fig:gt_creation}
\end{figure*}

The evolution of Neural Combinatorial Optimization (NCO) has been defined by a dialectic between \textit{expressiveness}—the ability to model multimodal posterior distributions—and \textit{tractability}—the computational cost of inference. While early learning-based approaches prioritized speed via constructive heuristics, the recent literature has shifted toward generative paradigms to close the optimality gap with classical solvers.

\subsection{Constructive and Autoregressive Baselines}
The foundational era of NCO treated optimization as a sequence generation task. Motivated by the rigidity of exact solvers, architectures such as Pointer Networks \cite{bello2016neural} and the Attention Model (AM) \cite{kool2018attention} parameterized the policy as an autoregressive encoder-decoder. These methods, typically trained via Reinforcement Learning or behavior cloning, offer highly efficient inference compared to search-based baselines. Subsequent works focused on stabilizing the high variance inherent in RL training; notably, POMO \cite{kwon2020pomo} leveraged the underlying symmetries of TSP to construct a low-variance baseline. 

However, constructive approaches suffer from a fundamental limitation: \textit{error propagation} inherent in greedy decoding. The sequential nature of decoding means that early, irreversible missteps permanently trap the model in local optima \cite{joshi2020learning}. While beam search offers marginal improvements, it scales poorly with problem size, failing to capture the global structural rearrangements required for high-quality solutions on large-scale instances.

\subsection{The Generative Shift: Diffusion and Consistency}
To overcome the limitations of autoregression, the field pivoted toward non-autoregressive, iterative refinement. DIFUSCO \cite{sun2023difusco} established a new paradigm by formulating TSP as a graph-based diffusion process, treating optimization as a heatmap denoising task. By modeling the joint probability of edges, these methods enable global refinement and capture multimodal optima, significantly reducing the optimality gap.

This expressiveness, however, incurs a severe inference bottleneck. Integrating the probability flow ODE often involves hundreds of function evaluations, creating a significant computational overhead. Recent "distilled" solvers have attempted to resolve this computational bottleneck. Fast-T2T \cite{li2024fast} employs consistency distillation to map noise directly to solutions in a single or few steps, while DISCO \cite{zhao2024disco} constrains the diffusion process to a "residue" manifold to drastically reduce the search space. 

\subsection{Geometric Decoupling and the Case for Flow}
While consistency models and residue constraints address latency, they inherit a critical structural limitation from their predecessors: the geometric decoupling of the solution space. Most state-of-the-art generative solvers (including DIFUSCO, Fast-T2T, and DIMES \cite{qiu2022dimes}) operate on the discrete $N \times N$ edge adjacency matrix.

This formulation approaches the problem indirectly. Recovering a structured geometric object (a permutation) from unstructured Gaussian noise on an edge manifold presents a challenging denoising task. This process focuses computational effort on resolving dense edge probabilities, rather than directly evolving the linear geometry of the tour itself.

CycFlow addresses this disconnect by adopting  a simulation-free alternative to diffusion that regresses a deterministic vector field, based on \textit{Flow Matching} \cite{lipman2022flow}. Unlike prior works that apply the flow to the edge space, we propose an instance-conditioned flow on the node coordinates. By transporting the input coordinates to a canonical circular arrangement, we align the generative process with the topological structure of the TSP, ensuring both linear-time tractability and geometric consistency.

\begin{table*}[t]
\caption{\textbf{TSP-50 and TSP-100 Results.} CycFlow matches the optimality of heavy iterative baselines while outperforming diffusion-based inference speeds by orders of magnitude. Baseline results are sourced from \cite{li2024fast}.}
\label{tab:results_tsp50_100}
\begin{center}
\begin{small}
\begin{sc}
\resizebox{0.75\textwidth}{!}{%
\begin{tabular}{l|cc|cc}
\toprule
& \multicolumn{2}{c|}{TSP-50} & \multicolumn{2}{c}{TSP-100} \\
Method & Gap (\%) & Time & Gap (\%) & Time \\
\midrule
\multicolumn{5}{l}{\textit{Exact \& Heuristics}} \\
Concorde & 0.00 & 3m & 0.00 & 12m \\
LKH3 \cite{helsgaun2017extension} & 0.00 & 3m & 0.00 & 33m \\
2Opt \cite{lin1973effective} & 2.95 & -- & 3.54 & -- \\
\midrule
\multicolumn{5}{l}{\textit{RL \& Constructive Baselines}} \\
AM \cite{kool2018attention} & 1.76 & 2s & 4.53 & 6s \\
GCN \cite{joshi2019efficient} & 3.10 & 55s & 8.38 & 6m \\
Transformer \cite{bresson2021transformer} & 0.31 & 14s & 1.42 & 5s \\
POMO \cite{kwon2020pomo} & 0.64 & 1s & 1.07 & 2s \\
Sym-NCO \cite{kim2022sym} & -- & -- & 0.94 & 2s \\
\midrule
\multicolumn{5}{l}{\textit{Diffusion \& Iterative Refinement}} \\
Image Diffusion \cite{graikos2022diffusion} & 1.23 & -- & 2.11 & -- \\
DIFUSCO (Speed) \cite{sun2023difusco} & 12.84 & 16s & 20.20 & 20s \\
DIFUSCO (Accuracy) \cite{sun2023difusco} & 0.41 & 18m & 1.16 & 18m \\
\midrule
\multicolumn{5}{l}{\textit{T2T \& Fast T2T (Recent SOTA)}} \\

T2T (Speed) \cite{li2023t2t} & 8.15 & 55s & 16.09 & 1m \\
T2T (Accuracy) \cite{li2023t2t} & 0.03 & 26m & 0.11 & 42m \\
Fast T2T (Speed) & 0.31 & 11s & 1.31 & 16s \\
Fast T2T (Accuracy) & \textbf{0.01} & 3m & \textbf{0.03} & 3m \\
\midrule
\textbf{CycFlow (Ours)} & 0.08 & \textbf{0.004s} & 0.35 & \textbf{0.01s} \\
\bottomrule
\end{tabular}
}
\end{sc}
\end{small}
\end{center}
\end{table*}

\section{Method}
\label{sec:method}
We address the Euclidean TSP problem, where the input is a collection of $N$ points in $\RR^2$, denoted by the point cloud $x_0\in \RR^{N\times 2}$ (throughout this paper, we use the terms ``point cloud,'' ``set of coordinates,'' and ``nodes'' interchangeably to refer to this input structure). The goal is to find the shortest Eulerian cycle which traverses through all the points. While this cycle was typically represented as a $N\times N $ permutation matrix in previous work,  a key innovation of our method is that we represent the optimal  cycle  geometrically  via another point cloud $x_1\in \RR^{N\times 2}$ whose entries all reside on a circle, forming a cycle. 

CycFlow solves TSP by transporting inputs $x_0\in \RR^{N\times 2}$ to a canonical target $x_1\in \RR^{N\times 2}$ via a learned ODE. This section details the data-dependent coupling that defines the boundary conditions (~\cref{subsec:coupling}), the conditional flow matching objective used to learn the dynamics (~\cref{subsec:flow_matching}), and our architectural choice to learn the velocity (\cref{subsec:architecture}).

\subsection{Data-Dependent Geometric Couplings}
\label{subsec:coupling}

\textbf{Target Construction.} Let $x_0 \in \mathbb{R}^{N \times 2}$ be the unordered input coordinates, and assume we know the optimal tour $\pi^*$ (which is the case in the training phase). We construct the target $x_1$ by embedding the optimal TSP tour onto a circle, scaled such that its Frobenius norm matches that of the input (i.e., $\|x_1\|_F = \|x_0\|_F$). The nodes are placed on this circle in the exact sequence of the optimal permutation $\pi^*$, with arc lengths strictly proportional to the edge weights of the original TSP coordinates relative to the full optimal cycle length (\cref{fig:gt_creation}). By explicitly constructing $x_1$ from the intrinsic geometry of $x_0$, we establish a deterministic relationship between the source and target.

\textbf{Coupling Alignment.} Our definition of $x_1$ still has a global rotation ambiguity. To minimize the complexity of the trajectory connecting $x_0$ and $x_1$, we perform Procrustes alignment. We explicitly seek the optimal rotation $R^*$ that minimizes the aggregate squared distance between each input node $(x_0)_i$ and its mapped target vertex $(x_1)_i$. This objective is mathematically equivalent to minimizing the squared Frobenius norm of the difference matrix:
\begin{equation}
\resizebox{\hsize}{!}{$
    R^* = \underset{R \in SO(2)}{\text{argmin}} \sum_{i=1}^N \| (x_1)_i R - (x_0)_i \|^2 \equiv \underset{R \in SO(2)}{\text{argmin}} \| x_1 R - x_0 \|_F^2
$}
\end{equation}
The minimizer $R^*$ is easily found using the Kabsch algorithm  \cite{Kabsch1978ADO}. The aligned pair $(x_0, x_1 R^*)$ constitutes a correlated sample from our joint distribution. This coupling strategy ensures that the global displacement is minimal, significantly reducing transport cost and preventing the formation of complex, high-curvature trajectories characteristic of independent couplings \cite{bose2024se3stochasticflowmatchingprotein}. By explicitly minimizing global displacement, this alignment acts as a structural "head start," simplifying the regression task from resolving long-range entanglements to learning local, low-velocity adjustments.

\subsection{Deterministic Conditional Flow Matching}
\label{subsec:flow_matching}

We reformulate the generation process as learning the conditional velocity field of a deterministic probability flow \cite{lipman2022flow}. We utilize the framework of \cite{albergo2023stochastic} in its noiseless limit, effectively constructing a direct transport map from the input to the solution.

\begin{figure}[t]
    \centering
    \includegraphics[width=\linewidth]{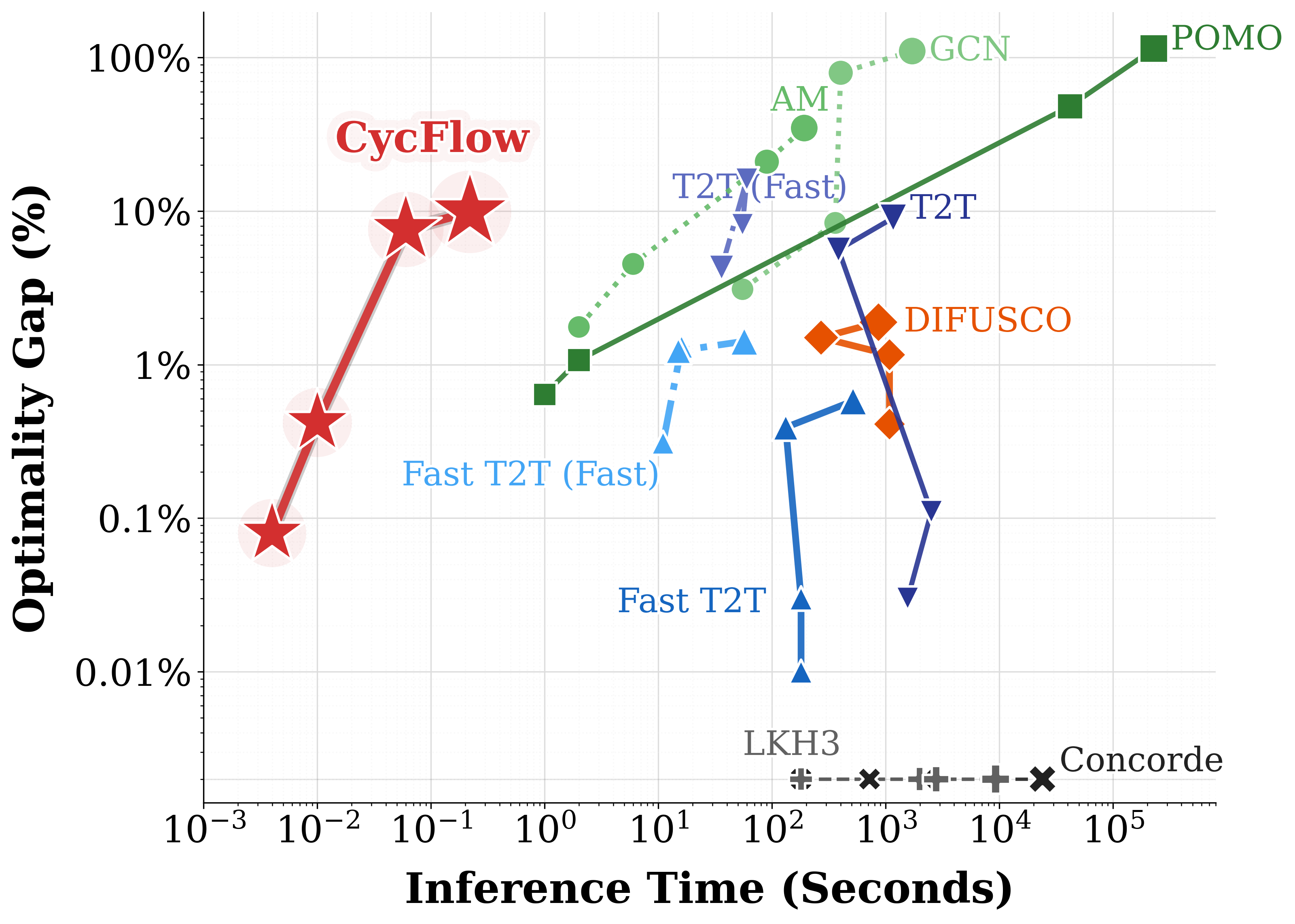}
    \caption{\textbf{Inference Latency vs.\ Optimality Gap (TSP-50 to TSP-1000).}
    We plot the time--accuracy Pareto frontier on a log--log scale (lower-left is better); marker sizes indicate problem size $N$.
    CycFlow (red stars) occupies a previously unreachable sub-second regime, achieving competitive optimality gaps while reducing inference latency by orders of magnitude compared to diffusion-based solvers (e.g., DIFUSCO), constructive methods (POMO), and exact solvers (Concorde).
    This demonstrates an expansion of the efficiency--accuracy Pareto frontier rather than a simple trade-off between speed and solution quality.}

    \label{fig:pareto_tsp}
\end{figure}

\textbf{The Linear Interpolant.}
We define a time-dependent trajectory $x_t$ that linearly interpolates between the unordered input coordinates $x_0$ at $t=0$ and the target manifold $x_1$ at $t=1$. Following the formulation of \cite{albergo2023stochastic} with the noise coefficient set to zero ($\gamma_t = 0$), the path is given by:
\begin{equation}
    x_t = (1-t)x_0 + t x_1, \quad t \in [0, 1].
\end{equation}
This construction leverages the input $x_0$ as a ``structurally informative prior'' for the target, rather than starting from uninformative Gaussian noise. This defines a direct bridge between the problem and its solution.

\textbf{Conditional Velocity Field.}
 The flow of this interpolant is governed by the simple Ordinary Differential Equation (ODE) 
\begin{equation}
    \frac{d}{dt} x_t =  u(x_0,x_1), \text{ where } u(x_0,x_1)=x_1-x_0.
 \end{equation}
This corresponds to the \textit{Optimal Transport ODE} (OT-ODE) limit of the Schrödinger Bridge \cite{liu20232}, where the dynamics become purely deterministic.

\textbf{Regression Objective.}
To learn the transport dynamics, we parameterize a conditional neural velocity field $v_\theta(t, x_t | x_0)$ to approximate the ground-truth flow. 

We minimize the flow matching objective \cite{lipman2022flow}, which regresses the neural velocity field $v_\theta$ onto this target vector field. This is strictly equivalent to a square loss regression over the data-dependent couples $(x_0, x_1)$ sampled from the joint distribution $\rho$ constructed in ~\cref{subsec:coupling}:
\begin{equation}\label{eq:loss}
    \mathcal{L}_{\text{CFM}}(\theta) = \mathbb{E}_{\substack{t \sim \mathcal{U}(0,1), \\ (x_0, x_1) \sim \rho}} \left[ \left\| v_\theta(t, x_t|x_0) - u(x_0,x_1) \right\|^2 \right]
\end{equation}
By minimizing this objective, $v_\theta$ learns to mimic the optimal linear path that pushes unordered nodes $x_0$ directly to their canonical solution arrangement $x_1$.

\textbf{Inference.}
At test time, given a new problem instance $x_0$, we generate the solution in two steps:
\begin{enumerate}
    \item \textbf{Integration:} We approximate the flow map  by numerically integrating the learned ODE starting at $X_0=x_0$ and using a simple Euler solver with $K$ steps:
    \begin{equation*}
        X_{k+1} = X_k + \frac{1}{K} v_\theta(t_k, X_k | X_0), \quad k=0 \dots K-1
    \end{equation*}
    This yields the transported points $ x_1:=X_k$, which lie on the target manifold (the canonical circle).
    
    \item \textbf{Recovery:} We recover the discrete permutation $\hat{\pi}$ by sorting the $N$ points in $x_1$  based on their polar angle relative to the origin. This solution is then refined using the local greedy 2OPT algorithm. 
\end{enumerate}

\paragraph{The Role of Couplings in Generative Frameworks.}

Standard generative models conventionally select a simple base density, such as a standard Gaussian, primarily for its analytical tractability and ease of sampling \cite{ho2020denoising,song2020score}. This results in an \textit{independent coupling}, where the starting distribution is agnostic to the target data structure, often forcing the model to learn complex trajectories to bridge two disparate distributions. In contrast, our approach leverages the framework of stochastic interpolants with data-dependent couplings \cite{liu20232, albergo2023stochastic, bose2024se3stochasticflowmatchingprotein}. Instead of treating the input and target as independent, we explicitly construct a \textit{dependent} pairing: for every unordered input set $x_0$, we generate a unique, corresponding target geometry $x_1$ derived directly from the optimal solution. Formally, this defines a sharp conditional distribution $\rho_1(x_1|x_0)$, creating a valid \textit{data-dependent coupling} $\rho(x_0, x_1) = \rho_0(x_0)\rho_1(x_1|x_0)$ that permits efficient learning via flow matching \cite{lipman2022flow}. 

\subsection{Architecture: Canonicalize-Process-Restore}
\label{subsec:architecture}
In this subsection we describe our architecture for the velocity field $v_\theta$. We design our architecture to be size-agnostic, allowing the same model to be applied to TSP problems of different sizes, and  strictly equivariant to permutation, rotation and translation of the input's pose. We attain equivariance using the  \textit{Canonicalize-Process-Restore} pipeline. This strategy decouples symmetry constraints from the network architecture. Unlike methods that rely on invariant aggregation—which often obscures fine-grained relative spatial structure—canonicalization allows us to utilize standard Transformers. This enables the model to remain ``geometrically aware,'' capturing precise relative node placement within a standardized frame while guaranteeing global equivariance.

\textbf{Canonicalization}
We employ a variant of the canonicalization scheme proposed by \citet{friedmann2025canonnet}.
To resolve permutation ambiguity, we induce a deterministic sequence order based on the intrinsic geometry of the point cloud. We construct a full graph with Gaussian kernel weights $W_{ij} = \exp(-\|x_i - x_j\|^2 / \sigma^2)$ and compute the normalized Graph Laplacian $L_{sym} = I - D^{-1/2} W D^{-1/2}$.
The points are sorted according to the values of the Fiedler vector (the eigenvector corresponding to the second smallest eigenvalue). To resolve the inherent sign ambiguity of this eigenvector $v$, we enforce positive skewness: if $\sum v_i^3 < 0$, we flip $v \leftarrow -v$. This ensures a consistent traversal direction across structurally similar instances.

Once ordered, we resolve rigid $E(2)$ ambiguities by fixing a canonical reference frame. 
First, we enforce translation invariance by centering the point cloud at the origin: $x_i \leftarrow x_i - \bar{x}$, where $\bar{x} = \frac{1}{N}\sum_{j=1}^N x_j$.
Next, we compute a weighted orientation vector $\mathbf{u} = \sum_{i=1}^N w_i \mathbf{x}_i$, 
where the weights $w_i$ vary linearly from $-1$ to $1$ along the spectral sequence. 
The point cloud is rotated such that $\mathbf{u}$ aligns with the positive vertical axis. 
Finally, reflection ambiguity is resolved by examining the latter half of the sum: 
we calculate the aggregate $x$-coordinate of the points in the second half of the sequence 
(where $w_i > 0$). If this sum is negative, we reflect the point cloud across the $y$-axis.

\begin{table*}[!t]
\caption{\textbf{TSP-500 and TSP-1000 Results.} Scalability comparison. CycFlow maintains sub-second inference on TSP-1000, whereas constructive baselines (POMO, DIMES) suffer from prohibitive runtime growth and iterative models (T2T) require minutes to converge. Baseline results are sourced from \cite{li2024fast}.}
\label{tab:results_tsp500_1000}
\begin{center}
\begin{small}
\begin{sc}
\resizebox{0.75\textwidth}{!}{%
\begin{tabular}{l|cc|cc}
\toprule
& \multicolumn{2}{c|}{TSP-500} & \multicolumn{2}{c}{TSP-1000} \\
Method & Gap (\%) & Time & Gap (\%) & Time \\
\midrule
\multicolumn{5}{l}{\textit{Exact \& Heuristics}} \\
Concorde & 0.00 & 37.7m & 0.00 & 6.65h \\
LKH3 \cite{helsgaun2017extension} & 0.00 & 46.3m & 0.00 & 2.57h \\
\midrule
\multicolumn{5}{l}{\textit{RL \& Constructive Baselines}} \\
AM \cite{kool2018attention} & 20.99 & 1.5m & 34.75 & 3.2m \\
GCN \cite{joshi2019efficient} & 79.61 & 6.7m & 110.29 & 28.5m \\
POMO \cite{kwon2020pomo} & 48.22 & 11.6h & 114.36 & 63.5h \\
DIMES \cite{qiu2022dimes} & 14.38 & 1.0m & 14.97 & 2.1m \\
\midrule
\multicolumn{5}{l}{\textit{Diffusion \& Iterative Refinement}} \\
DIFUSCO \cite{sun2023difusco} & 1.5 & 4.5m & 1.89 & 14.4m \\
\midrule
\multicolumn{5}{l}{\textit{T2T \& Fast T2T (Recent SOTA)}} \\

T2T (Speed) \cite{li2023t2t} & 4.28 & 36s & -- & -- \\
T2T (Accuracy) \cite{li2023t2t} & 5.61 & 6.4m & 9.04 & 19.4m \\
Fast T2T (Speed) & 1.23 & 15s & 1.42 & 57s \\
Fast T2T (Accuracy) & \textbf{0.39} & 2.2m & \textbf{0.58} & 8.6m \\
\midrule
\textbf{CycFlow (Ours)} & 6.84 & \textbf{0.06s} & 9.89 & \textbf{0.22s} \\
\bottomrule
\end{tabular}
}
\end{sc}
\end{small}
\end{center}
\end{table*}

\textbf{Processing and Restoration.}
The canonicalized sequence is fed into a Transformer backbone. We utilize Rotary Positional Embeddings (RoPE) to explicitly encode the relative positions of points along the spectral curve, allowing the attention mechanism to capture fine-grained geometric dependencies. Furthermore, the flow time $t$ is injected into the model via Adaptive Layer Normalization (AdaLN), which modulates the scale and shift parameters of each block based on the current timestep. The network predicts the velocity field $v_{\text{can}}$ in this standardized frame. Finally, we apply the inverse of the canonical rotation and permutation to transport the update back to the original input geometry.

\begin{table*}[t]
    \centering
	\begin{minipage}[t]{0.48\textwidth}
        \caption{\textbf{Impact of Flow Matching (TSP-50).} Comparison of our iterative transport against single-shot baselines. The \emph{Angular Sort} baseline shows that  zero-shot inference using our circular geometric prior yields meaningful but sub-optimal results. The failure of \emph{Direct Regression} demonstrates that a single-step prediction cannot resolve the complex point permutations, whereas our iterative ODE approach successfully closes the gap.}
        \label{tab:ablation_flow_vs_direct}
        \begin{center}
        \begin{small}
        \begin{sc}
        \renewcommand{\arraystretch}{1.55} 
        \begin{tabular}{l|c}
        \toprule
        Method & Gap (\%) \\
        \midrule
        Angular Sort & 9.38 \\
        Direct Angular Reg. & 3.75 \\
        \textbf{CycFlow (Ours)} & \textbf{0.09} \\
        \bottomrule
        \end{tabular}
        \end{sc}
        \end{small}
        \end{center}
    \end{minipage}
    \hfill
        \begin{minipage}[t]{0.48\textwidth}
        \caption{\textbf{Ablation on Backbone Architecture (TSP-50).} We compare our architecture against extensive hyperparameter sweeps of equivariant GNNs and Transformer variants. EGNNs struggle with fine geometry, while standard Transformers suffer from a lack of rotation equivariance. }
        \label{tab:ablation_backbone}
        \begin{center}
        \begin{small}
        \begin{sc}
        \begin{tabular}{l|c}
        \toprule
        Backbone & Gap (\%) \\
        \midrule
        Equivariant GNN & 0.34 \\
        Transformer & 0.20 \\
        Transformer + Fourier & 0.27 \\
        Transformer + ROPE & 0.18 \\
        \textbf{CycFlow (Ours)} & \textbf{0.09} \\
        \bottomrule
        \end{tabular}
        \end{sc}
        \end{small}
        \end{center}
    \end{minipage}
\end{table*}

\section{Experiments}
\label{sec:experiments}


We evaluate CycFlow on Euclidean TSP benchmarks with $N \in \{50, 100, 500, 1000\}$. Our main finding is that CycFlow achieves competitive accuracy with an inference time of several milliseconds, which is a $2-3$ order of magnitude speedup with respect to other diffusion baselines.

\subsection{Experimental Protocol}
We adhere to a rigorous protocol to ensure fair comparison of runtime and accuracy:

\textbf{Metrics.} We report the \textbf{Approximation Gap} relative to the  optimal tour length $L_{\text{opt}}$ (computed using Concorde \cite{applegate2009certification}):
\begin{equation}
    \text{Gap}(\%) = 100 \cdot \frac{L_{\text{method}} - L_{\text{opt}}}{L_{\text{opt}}}
\end{equation}
Runtime is reported as the median per-instance wall-clock time (in seconds), including the full inference stack (numerical integration, angular recovery, and parallel 2-opt refinement).

\textbf{Baselines.} We compare against a broad spectrum of solvers: exact methods (Concorde, LKH3), constructive heuristics (AM, POMO), and state-of-the-art iterative models (DIFUSCO, T2T, Fast-T2T). For baseline papers which reported results of several variants, we report both  ``best speed'' and ``best accuracy'' variants to expose the full trade-off profile. To ensure fairness, we compare against baseline checkpoints that utilize 2-opt post-processing whenever such results are available.

\textbf{Reproducibility.}  Code will be released with the final version of the paper.

\subsection{Main Results: The Latency-Accuracy Pareto Frontier}
Our results on smaller TSP problems with $N=50,100$ points are reported in 
~\cref{tab:results_tsp50_100}, and results for larger TSP instances $N=500, 1000$ are reported in  \cref{tab:results_tsp500_1000}. 

The two tables show that CycFlow achieves sub-second latency even on large-scale instances. For TSP-1000, our method's inference time is  \textbf{0.22 s}, whereas accuracy-focused diffusion baselines require minutes (e.g., Fast T2T requires 516 s). This represents a speedup of 2-3 orders of magnitude, validating that linear coordinate dynamics eliminate the quadratic bottleneck of edge-denoising. 

In terms of accuracy, on TSP-50 CycFlow attains a very low \textbf{0.08\%} gap in just \textbf{4 ms}, as the number of points $N$ increases we observe a more significant efficiency-accuracy tradeoff. On our largest instance,  TSP-1000, CycFlow yields a \textbf{9.89\%} gap at \textbf{0.22 s}. This accuracy is competitive with many previous methods, but is substantially higher than computationally intensive global solvers, as well as  Fast T2T's lower gap which is 2-3 orders of magnitude slower. Thus, CycFlow should  not be considered as a drop-in replacement for solvers where time is infinite and optimality is paramount; rather, it provides a viable neural solver for real-time applications requiring competitive solutions at scale.

 ~\cref{fig:pareto_tsp} gives another visualization of these results in the efficiency-accuracy plane. The figure shows that CycFlow is 'Pareto Optimal', in the sense that it is significantly faster than previous methods, while retaining competitive accuracy.




\subsection{Ablations and Analysis}
We conduct targeted ablations on TSP50 to isolate the source of our performance gains.

\textbf{Necessity of Flow Matching (\cref{tab:ablation_flow_vs_direct}).}
To isolate the contribution of the transport dynamics, we compared CycFlow against two single-shot baselines. We first evaluated a naive \emph{Angular Sort}, which simply orders the raw input points based on their polar angle. While this yields a high gap of 9.38\% on TSP50, it is far better than random, confirming that mapping points to a circle is a geometrically sound premise. Next, we tested \emph{Direct Angular Regression}, where the model attempts to predict the final target angles in a single forward pass. This method failed to generalize effectively (3.75\% gap), suggesting that the mapping from random 2D coordinates to a precise cycle is too complex and non-linear for a one-step prediction. CycFlow achieves a 0.09\% gap because the iterative ODE integration breaks this difficult transformation down into a continuous, smooth path, allowing the model to gradually steer points to their optimal positions.

\textbf{Backbone and Robustness (\cref{tab:ablation_backbone}).}
To isolate the effect of our overall framework and our specific choice of velocity model $v_\theta$ described in  \cref{subsec:architecture}, we experimented with replacing our architecture with other backbone architectures, conducting  extensive hyperparameter sweeps for each architecture to ensure a fair comparison. We first  evaluated Equivariant GNNs \cite{satorras2021n} on TSP50.  Despite their theoretical rotation invariance, they yielded the highest optimality gap (0.34\%). We hypothesize that the iterative message-passing paradigm may degrade the representation of the fine-grained global geometry required for high-precision TSP solving.

Next, we evaluated standard Transformer variants. While these models capture global interactions better than GNNs, they lack inherent rotation equivariance. This forces the model to expend capacity on learning approximate symmetries from the data, limiting its ability to focus purely on the combinatorial optimization task.

Ultimately, our cycFlow achieved the best performance (0.09\% gap) by employing the spectral canonicalization strategy described in  \cref{subsec:architecture}. We attribute the superiority of this method to two factors: first, canonicalization removes the burden of learning rigid symmetries, simplifying the geometric task. Second, the resulting spectral ordering assigns structural meaning to the sequence positions, allowing the model to effectively utilize standard Rotary Positional Embeddings (RoPE), where the specific frequencies naturally align with the spectral properties of the data.

\section{Conclusion and Limitation}

We have presented \textbf{CycFlow}, a framework that redefines the Euclidean TSP as a deterministic geometric transport problem rather than a stochastic heatmap generation task. By learning a continuous vector field that transports input coordinates to a canonical circular manifold, we replace the quadratic complexity of iterative edge scoring with \textbf{linear coordinate dynamics}. Empirically, this approach accelerates inference by orders of magnitude, with an inference time $\ll 1 \text{ sec}$ even for TSP1000, while maintaining competitive optimality gaps. CycFlow demonstrates that aligning the generative process with the problem's inherent geometry enables real-time, high-fidelity optimization.

\textbf{Limitations and Future Work.}
Our current framework operates within the supervised learning paradigm, requiring access to optimal solutions during training—a characteristic shared by leading diffusion-based NCO solvers \cite{sun2023difusco}. Developing unsupervised or reinforcement learning variants to eliminate this dependency remains an open frontier. Additionally, CycFlow is explicitly designed to exploit the continuous inductive bias of Euclidean space. Extending this transport-based methodology to non-Euclidean or general graph-based routing problems will require architectural adaptations to encode discrete edge metric structures in addition to geometric structure, representing a promising direction for future research in geometric flow matching. 

Ultimately, we believe that our fundamental ideas of using geometric flows, with a geometric cyclic representation of the optimal cycle, can be highly effective for fast TSP solving, in both supervised and unsupervised settings, and in the many TSP instances where Euclidean coordinates are informative, even if the distances are non-Euclidean. 

\section*{Impact Statement}

This paper presents work whose goal is to advance the field of Machine Learning. There are many potential societal consequences of our work, none which we feel must be specifically highlighted here.

\clearpage
\bibliography{example_paper}
\bibliographystyle{icml2026}

\end{document}